\documentclass[conference,a4paper]{IEEEtran}
\IEEEoverridecommandlockouts
\usepackage{cite}
\usepackage{caption}
\usepackage{subcaption}
\usepackage{array}
\usepackage{amsmath,amssymb,amsfonts}
\usepackage{algorithmic}
\usepackage{graphicx}
\usepackage{textcomp}
\usepackage{xcolor}
\usepackage{multirow}
\usepackage{booktabs,caption}
\usepackage{threeparttable}
\newcommand{\tabincell}[2]{\begin{tabular}{@{}#1@{}}#2\end{tabular}}
\def\BibTeX{{\rm B\kern-.05em{\sc i\kern-.025em b}\kern-.08em
    T\kern-.1667em\lower.7ex\hbox{E}\kern-.125emX}}

\IEEEoverridecommandlockouts
\IEEEpubid{\makebox[\columnwidth]{978-1-7281-8068-7/20/\$31.00~\copyright~2020 IEEE\hfill} \hspace{\columnsep}\makebox[\columnwidth]{}}

\begin{document}

\title{News Image Steganography: A Novel Architecture Facilitates the Fake News Identification\\}
\author{\IEEEauthorblockN{Jizhe Zhou}
\IEEEauthorblockA{\textit{Department of Computer and Information} \\
\textit{Science, University of Macao}\\
Macao, China \\
yb87409@um.edu.mo}
\and
\IEEEauthorblockN{Chi-Man Pun\IEEEauthorrefmark{1} }
\thanks{\IEEEauthorrefmark{1}Chi-Man Pun is the corresponding author}
\IEEEauthorblockA{\textit{Department of Computer and Information} \\
\textit{Science, University of Macao}\\
Macao, China \\
cmpun@umac.mo}
\and
\IEEEauthorblockN{Yu Tong}
\IEEEauthorblockA{\textit{vivo AI Lab} \\
\textit{vivo Mobile Communication Co., Ltd.}\\
ShenZhen, China \\
yb87462@umac.mo}
}

\maketitle

\begin{abstract}
A larger portion of fake news quotes untampered images from other sources with ulterior motives rather than conducting image forgery. Such elaborate engraftments keep the inconsistency between images and text reports stealthy, thereby, palm off the spurious for the genuine. This paper proposes an architecture named News Image Steganography (NIS) to reveal the aforementioned inconsistency through image steganography based on GAN. Extractive summarization about a news image is generated based on its source texts, and a learned steganographic algorithm encodes and decodes the summarization of the image in a manner that approaches perceptual invisibility. Once an encoded image is quoted, its source summarization can be decoded and further presented as the ground truth to verify the quoting news. The pairwise encoder and decoder endow images of the capability to carry along their imperceptible summarization. Our NIS reveals the underlying inconsistency, thereby, according to our experiments and investigations, contributes to the identification accuracy of fake news that engrafts untampered images.
\end{abstract}

\section{Introduction}
\begin{figure*}[htbp]
\centerline{\includegraphics[scale=0.28]{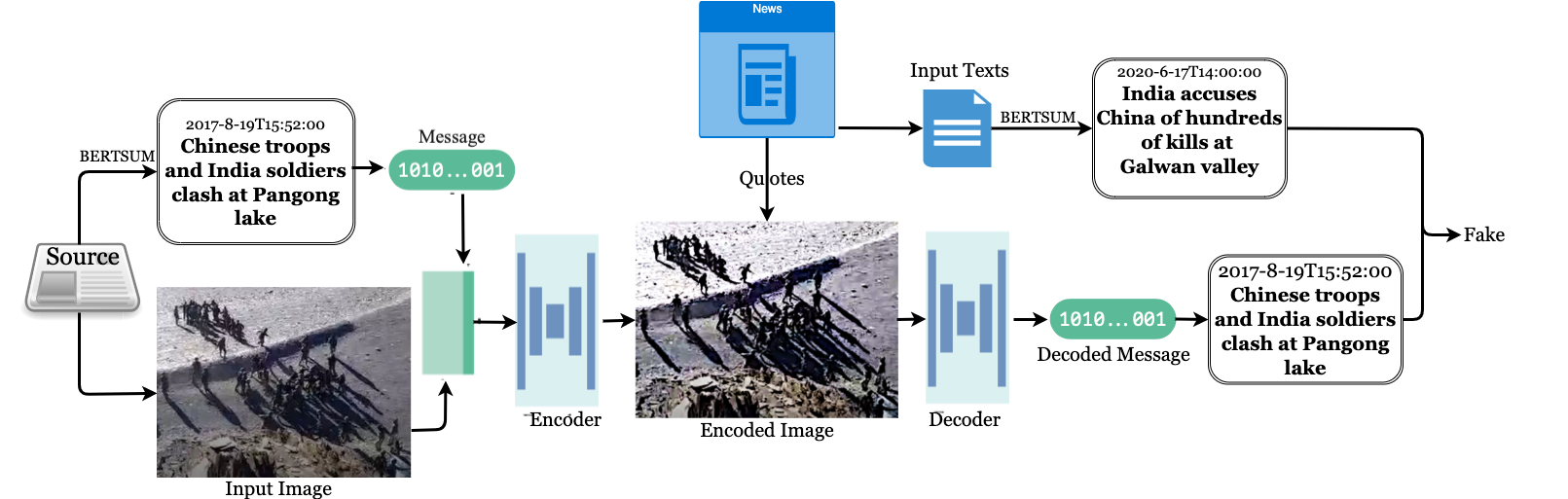}}
\caption{News Image Steganography (NIS) architecture to identify fake news according to image steganography. The summarization of an image is generated by BERTSUM~\cite{b8}. The encoder network then processes the input image and the summarization into the encoded image. When news quotes the encoded image, the decoder network recovers the summarization bitstring from the encoded image and translates it into summarization. The recovered summarization is leveraged as the ground truth to facilitate the identification of fake news.}
\label{fig1}
\vspace{-1em}
\end{figure*}

\par
Fake news detection or recognition is mainly conducted unilaterally either on text reports through Natural Language Processing (NLP) or on images through forgery detection~\cite{b1,b2}. As each part is equivalent in supporting the other mutually, state-of-the-art studies apply multimodal representation bilaterally~\cite{b3}. Therefore, both images forgery and reports integrity are valued in the assessment. However, there still exists another form of fake news that quotes untampered images from other sources maliciously to polish its statement and mislead the readers~\cite{b1}. Besides evading the forgery detection~\cite{b4}, such image engraftments deceive the multimodal learning networks and readers into trusting the news even more.
\par
The direct reason for current models incapability is the engrafted images cover the inconsistency between fake text reports and images~\cite{b5}. Multimodal representation or fusion aims to reveal the inconsistency between quoted images and the texts, thereby identify fake news among real ones~\cite{b6}. Limited by current techniques on image recognition and understanding, image engraftments along with the elaborately made-up texts keep the inconsistency stealthy to networks. Footprints in the sand show where one has been. Image engraftments adopt images that already exist but describe other real events. Thus, we also can easily identify these malicious engraftments as long as we know the source description or report of the image.
\par
Revelated by the information hiding in traditional image forgery detection~\cite{b7}, we propose a novel architecture named News Image Steganography (NIS) to reveal the stealthy inconsistency by allowing the images to carry their own summarization along. The summarization of an image can be yielded by NLP algorithms based on its reliable source news~\cite{b8}. Leveraging the steganographic algorithm, summarization about an image can be hidden within itself invisibly. As the hidden information is also desired to be presented once needed, the steganographic algorithm shall be constructed in a dual encoder-decoder structure. Therefore, Generative Adversarial Networks (GAN)~\cite{b9} is the best choice. When an encoded image is quoted, the corresponding decoder can detect and extract the imperceptible summarization from the image. Summarization of the quoted image can be presented to readers as the ground truth for the verification. Or we can generate summarization of the quoting news through the same algorithm and further compare it with the image's. Our experiments show NIS architecture preserves the summarization of an image after decoding; gains robustness both on digital and paper prints; and significantly contributes to the identification accuracy of the news with image engraftments.
\par
In general, the contribution of our paper contains:
\begin{itemize}
\item We propose a brand-new News Image Steganography (NIS) architecture based on information hiding. NIS facilitates and contributes to the identification accuracy of fake news that grafts untampered images maliciously.
\item We establish a GAN-based image steganographic algorithm in NIS to encode the summarization of an image within itself invisibly, and decode the summarization lossless regardless of the media used to present the image.
\end{itemize}

\section{Related Works}
\subsection{Steganography}
Image steganography is the algorithm of concealing messages within an image, especially in an invisible manner. Current studies of steganography mainly are developed on the GAN by training pairwise encoding and decoding networks. Deep Steganography~\cite{b10} is the pioneer of this area. They build their steganographic algorithm on GAN with minimal perturbations. Their results are acceptable in idea environments. However, with minimal perturbations in training, Deep Steganography~\cite{b10} cannot reach satisfying performance under a close-real environment. HiDDeN~\cite{b11} is more stable under a close-real environment due to pixelwise perturbations and random cropping are applied. These two data augmentation methods make HiDDeN~\cite{b11} noise and fouling resistant, but still prone to compression and perspective changes in printing. LFM~\cite{b12} introduces the light field messages to establish an extra spatial perturbation in training. Therefore, LFM~\cite{b12} is effective in resisting the perspective and compress changes in recaptured images. StegaStamp~\cite{b13} is state-of-the-art work on steganography which aims to hide hyperlinks into images. StegaStamp~\cite{b13} considers the real-world environment and introduces both pixelwise and spatial perturbations in the training stage to produce an encoder-decoder pair that is robust to noises, perspective changes, distortions, and compression. In this paper, we adopt the StegaStamp~\cite{b13} as the backbone in building our NIS network that embeds the summarization into its image secretly.
\subsection{Fake News Detection}
Fake news detection, recognition or identification is a hot topic in NLP community rather than the community of computer vision and image processing. Tons of works try to identify fake news based on the texts of the news~\cite{b1,b2,b4}. Surely the texts carry the major information, however, if we check the other side of the coin, images in the news just play an as pivotal role as the texts. Reliable news will use images to amplify its persuasions. As based on voids, fake news even prefers to quote or tamper images more. In such a way, images can polish its fake statements to mislead the readers for malicious goals. Some latest work use forgery detection~\cite{b7} on images and multimodal representation learning~\cite{b5,b6} to identify the fake news that uses tampered images. However, these are still no solution for news that quotes untampered images for malicious use. Our NIS architecture resolves the identification problem as an information hiding, recovering and verification process and offers a brand-new way to solve the problem through image steganography techniques.

\section{Proposed Methods}
\subsection{Extractive Summarization}
Our general working pipeline is depicted in Fig~\ref{fig1}. For the commencement, we generate the extractive summarization for the images based on BERT~\cite{b14} with given reliable source texts. BERT is a pre-trained transformer model and gains tremendous success in most fields of the Natural Language Process (NLP). BERTSUM~\cite{b8} is the extended model of BERT that achieves state-of-the-art performance in the extractive summarization task. The extractive summarization is a short version that retains the key information and abstraction of the entire text or known as the predigested news. We intentionally apply the extractive summarization instead of the abstractive summarization which demands generative models and trends to yield unpredictable results for texts in-the-wild.
\par
We tune the pre-trained model of BERTSUM~\cite{b8} on the CNN/NYT daily mail datasets~\cite{b15} to ensure the date and time information about the source news is contained in summarization. Rogue score~\cite{b16} is the criterion for the greedy algorithm in selecting the summarization out from the candidates. In addition, the length of the summarization for a single image is strictly limited to 128 bits according to the capacity of our steganographic algorithm.

\subsection{Image Steganography}
Data transmission by information hiding has long been studied in both the steganography and watermarking fields. With the prevailing of deep neural networks, the Generative Adversarial Networks (GAN)~\cite{b9} is also leveraged to develop robust end-to-end image steganography. Moreover, considering the media of news, the encoder and decoder shall achieve robust performance even for physically printed images. In this paper, we adopt the StegaStamp~\cite{b13} as the backbone in developing our news image steganography network. StegaStamp~\cite{b13} introduces extra spatial image corruptions between the encoder and decoder in the GAN based framework. During the training, the spatial image corruptions analogize noises and approximate the space of distortions. Therefore, StegaStamp~\cite{b13} is robust in retrieving encoded bits in real-world conditions while preserving excellent perceptual image quality.
\par
Based on StegaStamp~\cite{b13}, we use an U-NET structure for our encoder. Three input channels are 400*400 pixel and an extra channel of the same size is intended for the summarization binary bitstring. Fully connected convolution layers are applied to the bitstring channel as we found that these layers boosts the convergence in the learning stage.
\par
In addition to pixelwise perturbations, we employ the spatial transformer network~\cite{b17} in the decoder to enhance its resistance to the perspective changes. Small perspective changes including rotations and distortions are common in printing. Therefore, spatial corruptions is the key in developing high robustness against the printed news media like newspapers. Since the summarization channel is processed independently from the other three image channels, a series of convolutional transformation are imposed on the images to ensure the unanimous output length for all channels. The cross-entropy loss is applied in supervising the convergence of the decoder.

\par
Similar to the StegaStamp~\cite{b13}, in the loss fucntion, $L_2$ regularization factor is denoted as $L_R$. $L_2$ penalty is effective in pulling the learning model back from the distracting patterns. Slight increases on the weight of $L_R$ can achieve so. $L_P$ is the LPIPS~\cite{b19} perceptual loss. The critic loss in StegStamp is discarded as the detection task is not part of our goal. The critic loss belongs to the training of the critical network which predicts whether a summarization is embedded within an image. $L_M$ is the cross-entropy loss for the decoded summarization bitstring. The total loss is the weighted sum of the aforementioned loss. $\lambda$ is the weight for the corresponding loss. The weights for three image channels' loss $\lambda_R, \lambda_P$ are initialed to zero for better accuracy of the decoder. Training is conducted on the Kaggle dataset of fake news~\cite{b18}.

\begin{equation}
L =	\lambda_R L_R +\lambda_P L_P+\lambda_M L_M
\end{equation}

\section{Experiments and Results}

\subsection{Steganography Experiments and Results}
Experiments are conducted on the fake news dataset offered by Kaggle~\cite{b18}. The dataset collects the labeled fake news and corresponding reliable source news for proof. We manually select 2000 pairs of fake/reliable source news that use the same untampered images for testing. Moreover, 100 out of these 2000 images are printed by different printers to simulate the images printed in newspapers or magazines in order to evaluate the performance of the decoder in the real-world. The decoding network is wrapped and migrated into a cellphone App in advance. Such that we can recapture the printed image by the cellphone camera and decode the summarization through the wrapped decoder.
\par
The images are printed using two consumer printers: HP LaserJet MFP M436 Printer and HP Neverstop Laser MFP 1202w. Also, the images are resized into different resolutions range from 200*200 to 720*720 mimicking the operations may happen in real situations. The recovered summarization results are listed in the TABLE~\ref{tbl}.
\begin{table}[htbp]
\centering
\caption{Accuracy of recovered summarization bitstring}
\begin{tabular}{|c|c|c|c|c|c|}
\hline
\tabincell{c}{Methods} &{Printer} &\tabincell{c}{5th } &\tabincell{c}{25th }  &\tabincell{c}{50th}   &Mean    \\
\hline\hline
\multirow{2}{*}{\tabincell{c}{Ours on \\ Printed Images}}
& \tabincell{c}{HP LaserJet\\ MFP M436}  & {95\%} & {98\%}  & {100\%} & {99\%}  \\
\cline{2-6}
& \tabincell{c}{HP Laser \\MFP 1202w} & {94.5\%} & {98\%}  & {99\%} & {98\%}  \\
\cline{2-6}
\hline\hline
{\tabincell{c}{Ours on\\ Digital Images}}
& N/A  & {N/A} & {N/A}  & {N/A} & {99.8\%}  \\
\cline{2-6}
\hline\hline
\cline{2-6}
\hline\hline
\multirow{2}{*}{\tabincell{c}{HiDDeN~\cite{b11} on \\Printed Images}}
& \tabincell{c}{HP LaserJet\\ MFP M436}  & {67.1\%} & {68.7\%}  & {69\%} & {68.5\%}  \\
\cline{2-6}
& \tabincell{c}{HP Laser \\MFP 1202w} & {66.3\%} & {67\%}  & {68.2\%} & {68\%}  \\
\cline{2-6}
\hline\hline
{\tabincell{c}{HiDDeN~\cite{b11} on \\Digital Images}}
& N/A  & {N/A} & {N/A}  & {N/A} & {99.2\%}  \\
\hline\hline
\cline{2-6}
\hline\hline
\multirow{2}{*}{\tabincell{c}{LFM~\cite{b12} on \\Printed Images}}
& \tabincell{c}{HP LaserJet\\ MFP M436}  & {90.4\%} & {92\%}  & {93\%} & {92\%}  \\
\cline{2-6}
& \tabincell{c}{HP Laser \\MFP 1202w} & {89.8\%} & {91\%}  & {93\%} & {91.6\%}  \\
\cline{2-6}
\hline\hline
{\tabincell{c}{LFM~\cite{b12} on \\Digital Images}}
& N/A  & {N/A} & {N/A}  & {N/A} & {99.6\%}  \\
\cline{2-6}

\hline
\end{tabular}
\begin{tablenotes}
\small
\item The decoding accuracy of the summarization bitstring (percentage of bits correctly recovered). The 100 images randomly chosen from the 2000 images are printed and then decoded.
\end{tablenotes}
\vspace {-1em}
\label{tbl}
\end{table}

\begin{table*}[htbp]
\centering
\caption{Questionnair Results}
\begin{tabular}{|c|c|c|c|c|c|c|}
\hline
& \tabincell{c}{Number of\\News} & \tabincell{c}{Number of\\ Fake News}  & \tabincell{c}{Number of\\Reliable News}& \tabincell{c}{Marked as Fake/\\ Fake News} &\tabincell{c}{Marked as Fake/\\ Reliable News} & \tabincell{c}{Total \\Accuracy}\\
\cline{2-7}
\hline\hline
{\tabincell{c}{First\\Round}}
& 516  & 271 & 235  & \tabincell{c}{94/271\\(34.6\%)} & \tabincell{c}{102/235\\(43.4\%)} &{18.2\%} \\
\cline{2-7}
\hline\hline
{\tabincell{c}{Second\\Round}}
& 516  & 271 & 235  & \tabincell{c}{181/271\\(66.8\%)} & \tabincell{c}{66/235\\(28.1\%)} &{35.1\%} \\
\cline{2-7}

\hline
\end{tabular}
\begin{tablenotes}
\small
\item The total accuracy and the recall rate of fake news are both nearly doubled in the second round. This indicates our architecture that encodes and decodes reliable summarizations from images heavily contributes to the identification process of fake news.
\end{tablenotes}
\vspace {-1em}
\label{tbl1}
\end{table*}
\par
From TABLE~\ref{tbl}, NIS is highly robust regardless of printers. The 5th, 25th, and 50th percentiles and mean of the printed images are shown as they are randomly sampled from the entire images dataset. The 5th percentile accuracy indicates an at least 94.5\% decoding accuracy. Our mean accuracy over all 100 printed images is 98.5\% considering both printers. For the digital media that quotes digital images, all 2000 images are tested by feeding to the decoding network directly, and the mean accuracy is 99.8\% at the margin. Similarly, we also build a lateral comparison among widely accepted steganographic algorithms including HiDDeN~\cite{b11}, LFM~\cite{b12}, and our method. Our network gains surpassing performances particularly on printed images. This performance helps our architecture to function well on printed media like newspapers, leaflets, and handbills.

\subsection{News Identification Experiments and Results}
After successfully decoding and recovering the carried summarization in the image, we distributed questionnaires to investigate whether the source summarization of the quoted image actually facilitates to identify the fake news. We create each web questionnaire by randomly selecting three news from the Kaggle dataset~\cite{b18}. At least one news is selected under the fake label and at least one news is selected under the reliable source label. Participants are required to read all three news without extra information in the first round. The exact number of fake news in a certain questionnaire is not known to the participants, and they are going to mark the fake ones they think. Afterward, the decoded summarization is offered for images quoted in the news. The second round of news identification is conducted with this information to test the efficacy of our NIS architecture.
\par
\begin{figure}[h]
\begin{center}$
\begin{array}{cc}
\includegraphics[width=38mm]{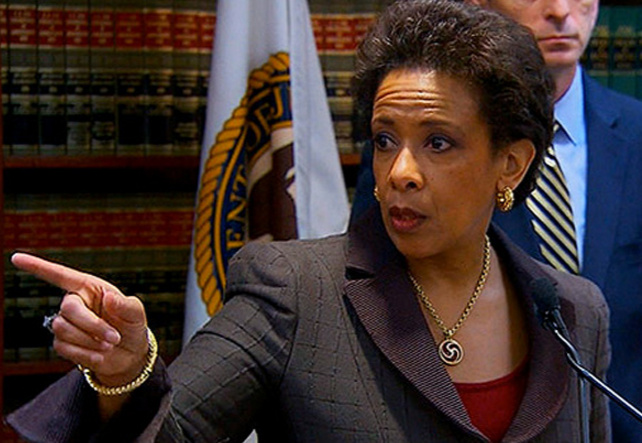}& \hspace{-0.8em}
\includegraphics[width=38mm]{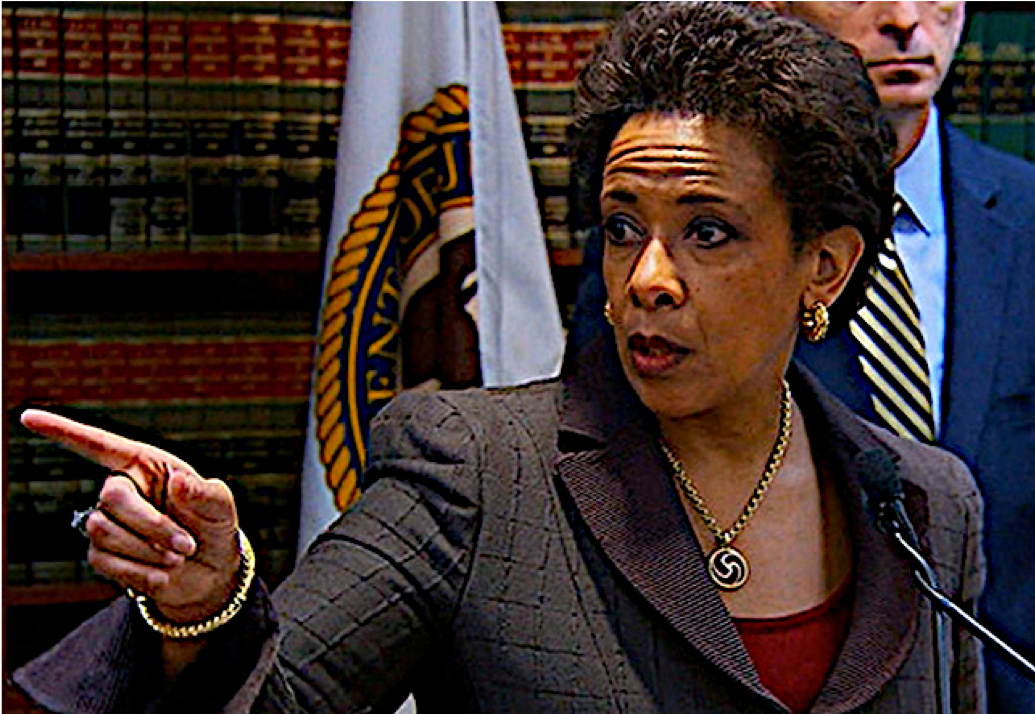} \hspace{-0.8em}\\
\vspace{-0.2em}
\includegraphics[width=38mm]{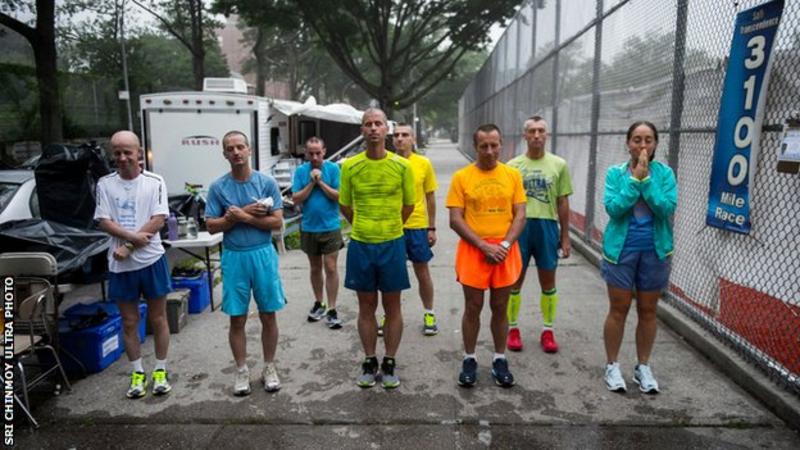}& \hspace{-0.8em}
\includegraphics[width=38mm]{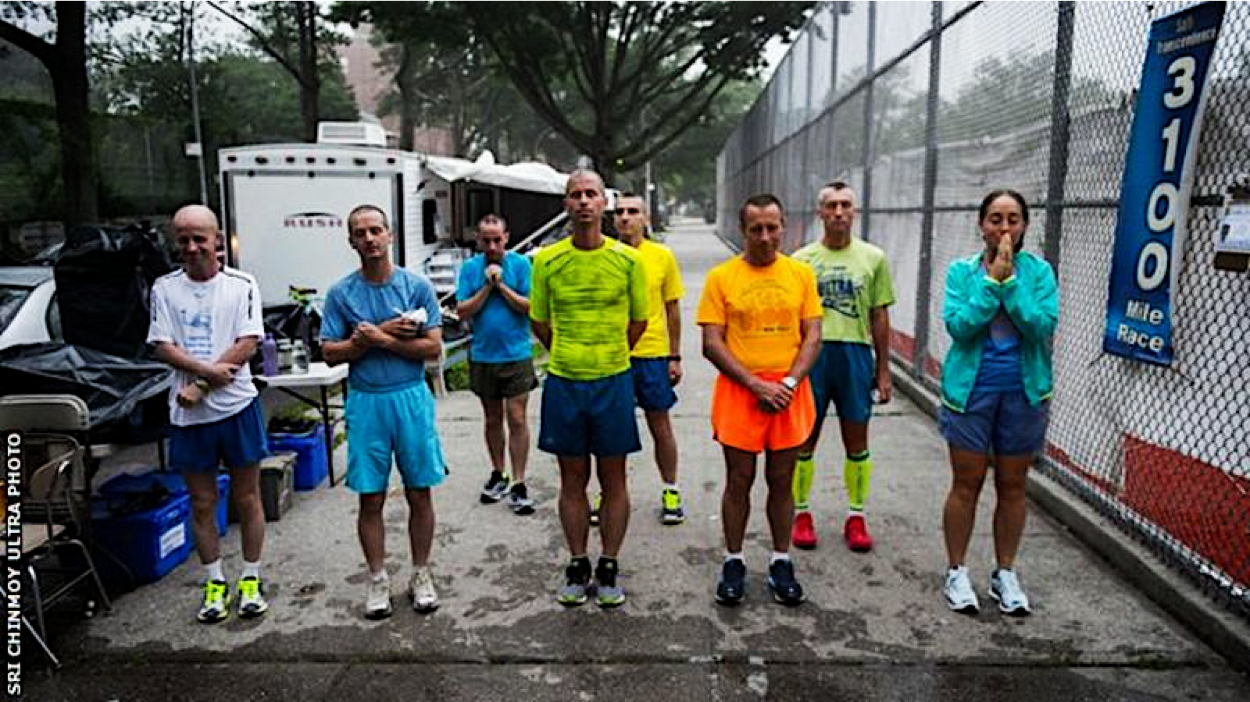} \hspace{-0.8em}\\
\vspace{-0.2em}
\end{array}$
\end{center}
\vspace{-0.8em}
\caption{The original image in the reliable source (left) and the corresponding encoded image ready to be quoted (right). Our NIS hides the summarization in an imperceptible manner.}
\label{fig2}

\end{figure}

We have received more than 200 feedback of the questionnaires, and 172 of them are full-filled and valid. The table~\ref{tbl1} indicates the overall efficacy of our NIS architecture. Obviously, NIS significantly improves the readers' judgment on news reliability. Both the recall rate and accuracy in fake news identification are nearly doubled in the second round. Thus, NIS enormously contributes to the fake news identification accuracy via image steganography. Some more results of encoded image are shown in the Fig~\ref{fig2}.

\section{Conclusion}
In this paper, we build the GAN-based News Image Steganography (NIS) architecture to endow the news images of the ability to carry along their own summarization. The experiments on news indicate NIS is capable of facilitating the identification of fake news quoting untampered images. We hope to promote this NIS architecture in daily applications and news source databases. If all news images can be encoded by our NIS, perhaps, the fake news has nowhere to hide.
\section{Acknowledgment}
This work was partly supported by the University of Macau under Grants: MYRG2018-00035-FST and MYRG2019-00086-FST, and the Science and Technology Development Fund, Macau SAR (File no. 0034/2019/AMJ, 0019/2019/A).


\begin{thebibliography}{00}
\bibitem{b1} Conroy, Nadia K., Victoria L. Rubin, and Yimin Chen. "Automatic deception detection: Methods for finding fake news." Proceedings of the Association for Information Science and Technology 52.1 (2015): 1-4.
\bibitem{b2} Shu, Kai, et al. "Fake news detection on social media: A data mining perspective." ACM SIGKDD explorations newsletter 19.1 (2017): 22-36.

\bibitem{b3} Ruchansky, Natali, Sungyong Seo, and Yan Liu. "Csi: A hybrid deep model for fake news detection." Proceedings of the 2017 ACM on Conference on Information and Knowledge Management. 2017.

\bibitem{b4} Huh, Minyoung, et al. "Fighting fake news: Image splice detection via learned self-consistency." Proceedings of the European Conference on Computer Vision (ECCV). 2018.


\bibitem{b5} Parikh, Shivam B., and Pradeep K. Atrey. "Media-rich fake news detection: A survey." 2018 IEEE Conference on Multimedia Information Processing and Retrieval (MIPR). IEEE, 2018.



\bibitem{b6} Khattar, Dhruv, et al. "Mvae: Multimodal variational autoencoder for fake news detection." The World Wide Web Conference. 2019.


\bibitem{b7} Mahdian, Babak, and Stanislav Saic. "A bibliography on blind methods for identifying image forgery." Signal Processing: Image Communication 25.6 (2010): 389-399.

\bibitem{b8} Liu, Yang. "Fine-tune BERT for extractive summarization." arXiv preprint arXiv:1903.10318 (2019).

\bibitem{b9} Radford, Alec, Luke Metz, and Soumith Chintala. "Unsupervised representation learning with deep convolutional generative adversarial networks." arXiv preprint arXiv:1511.06434 (2015).

\bibitem{b10} Baluja, Shumeet. "Hiding images in plain sight: Deep steganography." Advances in Neural Information Processing Systems. 2017.



\bibitem{b11} Zhu, Jiren, et al. "Hidden: Hiding data with deep networks." Proceedings of the European conference on computer vision (ECCV). 2018.


\bibitem{b12} Wengrowski, Eric, and Kristin Dana. "Light field messaging with deep photographic steganography." Proceedings of the IEEE Conference on Computer Vision and Pattern Recognition. 2019.
\bibitem{b13} Tancik, Matthew, Ben Mildenhall, and Ren Ng. "StegaStamp~\cite{b13}: Invisible hyperlinks in physical photographs." Proceedings of the IEEE/CVF Conference on Computer Vision and Pattern Recognition. 2020.
\bibitem{b14} Devlin, Jacob, et al. "Bert: Pre-training of deep bidirectional transformers for language understanding." arXiv preprint arXiv:1810.04805 (2018).

\bibitem{b15} Chen, Danqi, Jason Bolton, and Christopher D. Manning. "A thorough examination of the cnn/daily mail reading comprehension task." arXiv preprint arXiv:1606.02858 (2016).


\bibitem{b16} Lin, Chin-Yew. "Rouge: A package for automatic evaluation of summaries." Text summarization branches out. 2004.

\bibitem{b17} Jaderberg, Max, Karen Simonyan, and Andrew Zisserman. "Spatial transformer networks." Advances in neural information processing systems. 2015.


\bibitem{b18} Meg Risdal. "Getting Real about Fake News: Text and metadata from fake and biased news sources around the web" Dec. 2016. Accessed on: April. 2, 2020. [Online]. Available: https://www.Kaggle dataset~\cite{b18}.com/mrisdal/fake-news/data

\bibitem{b19} Zhang, Richard, et al. "The unreasonable effectiveness of deep features as a perceptual metric." Proceedings of the IEEE conference on computer vision and pattern recognition. 2018.

\end{thebibliography}
\end{document}